\documentclass{article}

 \PassOptionsToPackage{numbers, compress}{natbib}
\usepackage[final]{nips_2017}
\usepackage{setspace}
\usepackage[utf8]{inputenc} 
\usepackage[T1]{fontenc}    
\usepackage{hyperref}       
\usepackage{url}            
\usepackage{booktabs}       
\usepackage{amsfonts}       
\usepackage{nicefrac}       
\usepackage{microtype}      
\usepackage{graphicx}
\title{Denoising Arterial Spin Labeling Cerebral Blood Flow Images Using Deep Learning}

\author{
  Danfeng Xie \\
  Temple University\\
  Philadelphia, PA 19122 \\
  \texttt{danfeng.xie@temple.edu} \\
  \And
    Bai Li\\
   Temple University\\
   Philadelphia, PA 19122 \\
   \texttt{lbai@temple.edu} \\
   \AND
  Ze Wang \\
   Temple University\\
   Philadelphia, PA 19122 \\
   \texttt{ze.wang@temple.edu} \\
}

\begin{document}
\maketitle

\begin{abstract}
  Arterial spin labeling perfusion MRI is a noninvasive technique for measuring quantitative cerebral blood flow (CBF), but the measurement is subject to a low signal-to-noise-ratio(SNR). Various post-processing methods have been proposed to denoise ASL MRI but only provide moderate improvement. Deep learning (DL) is an emerging technique that can learn the most representative signal from data without prior modeling which can be highly complex and analytically indescribable. The purpose of this study was to assess whether the record breaking performance of DL can be translated into ASL MRI denoising. We used convolutional neural network (CNN) to build the DL ASL denosing model (DL-ASL) to inherently consider the inter-voxel correlations. To better guide DL-ASL training, we incorporated prior knowledge about ASL MRI: the structural similarity between ASL CBF map and grey matter probability map. A relatively large sample data were used to train the model which was subsequently applied to a new set of data for testing. Experimental results showed that DL-ASL achieved state-of-the-art denoising performance for ASL MRI as compared to current routine methods in terms of higher SNR, keeping CBF quantification quality while shorten the acquisition time by 75\%, and automatic partial volume correction.
\end{abstract}

\section{Introduction}
Arterial spin labeling (ASL) perfusion MRI is a technique for measuring cerebral blood flow (CBF) \cite{detre1992perfusion,williams1992magnetic}. In ASL, arterial blood water is labeled with radio-frequency (RF) pulses in locations proximal to the tissue of interest, and perfusion is determined by pair-wise comparison with separate images acquired with control labeling using various subtraction approaches \cite{aguirre2002experimental,lu2006detrimental,liu2005signal}. Limited by the longitudinal relaxation rate (T1) of blood water and the post-labeling transmit process, only a small fraction of tissue water can be labeled, resulting in a very low SNR \cite{wong1999potential}.

To improve SNR, a series of ASL images are usually acquired to take the mean perfusion map for final CBF quantification. Also, various preprocessing and analysis methods such as motion correction and outlier cleaning, have been proposed to further denoise ASL MRI \cite{wang2008empirical,detre2012applications}. However, those methods typically suffer from two major disadvantages. First, due to very poor original image quality, those methods achieve relative SNR improvement. Second, those methods usually involve a optimization process in the testing stage, which is very time-consuming.

Recenlty, Deep learning-based denoising methods emerged and achieves state-of-the-art performance \cite{mao2016image,zhang2017beyond,DBLP:journals/corr/LiuF17}.Instead of modeling explicit image prior, deep learning-based image denoising method learns image prior implicitly. The reasons of using CNN for denoising are as follows: First, CNN with deep or wide architecture \cite{zhang2017beyond} has the capacity and flexibility to effectively learning the image prior. Second, various well-developed training strategies and techniques are available to fasten the training process and improve the denoising performance, such as Rectifier Linear Unit (ReLU) \cite{nair2010rectified}, dropout \cite{srivastava2014dropout}, residual learning \cite{he2016deep} and batch normalization \cite{ioffe2015batch}. Third, CNN can be trained on modern powerful GPU using parallel computation, which  further improve the run time performance.

To the best of our knowledge, this work is the first deep learning-based method for denoising ASL perfusion MRI images. The purpose of this study was to assess the feasibility and efficacy of deep learning-based method for denoising ASL perfusion MRI images. For ease of description, we dubbed the new DL-based ASL denoising method as ASLDLD thereafter.

Our model, ASLDLD, was based on Wide Inference Network (WIN) \cite{DBLP:journals/corr/LiuF17}, a 5 layer end-to-end Convolutional Neural Networks (CNNs) denoising model with wide structure.  The ASLDLD was trained on 3D ASL MRI images acuqired from 280 subjects, where each subject has 20 slices and each slice has 40 control/labeled image pairs. The ASLDLD takes mean of first 10 CBF images without smoothing (meanCBF-10) as the input noisy image and the mean of all 40 CBF images (meanCBF-40) with smoothing and adaptive outlier cleaning \cite{wang2013arterial} as the reference image.  In order to fasten and stabilize model learning, we adopt residual learning and batch normalization learning strategies. Furthermore, Grey matter (GM) probability map was incorporated as a regularizer because CBF map shows a similar image contrast to that of a grey matter map.

ASLDLD has several advantages in denoising ASL MRI images: 1) the model effectively utilized prior information which significantly improve the denoising performance. 2) Because of the intrinsic of feed-forward CNN architecture, the computaion time is very fast in the test stage, which significantly reduce the computation time. (Contrast to traditional denoising method, which requires very long time to computing.) 3) Comparing traditional ASL denoising methods which requires a large series of label controling image pairs (in our case, 40 pairs), ASLDLD only need 10 pairs of label controlling images. This significantly reduces the acquisition time of ASL MRI, which reduce the chance of head motions and hence reduce the chance of introducing extra noise.

The rest of this paper is as follows. In section 2, we discuss about the related works of deep learning-based denoising methods. In section 3, we present the proposed ASLDLD architecture. Section 4 demonstrates the experiment of our methods on data. Last but not least, in section 5, we discuss the main contributions and results of this work.

\section{Related Works}

Imaging denoising is a classic low-level vision problem which have been widely studied in past decades.
The image prior modeling often play a central role in image denoising. Traditional methods that used image prior knowledge as regularization techniques, scuh as nonlocal self-similarity models \cite{buades2005non,dabov2007image,buades2008nonlocal,mairal2009non}, Markov Random Field (MRF)\cite{lan2006efficient,li2009markov,roth2009fields} and spares models \cite{elad2006image,dong2013nonlocally}, have shown very promising performance. However, in traditional denoising methods, image prior knowledge are explicitly pre-defined, which are often limited in capturing the full characteristics of image structure and limited in blind image denoising.

Neural network based denoising method is another active category of image denoising. The main difference between the neural network based methods and other methods is that neural network typical learn image prior implicitly rather than pre-defined image priors by training directly on pairs of noisy images and corrupted images. The most popular neural network based denoising methods up-to-date are based on Convolutional neural networks(CNNs). CNNs learn a hierarchy of features by a series of convolution, pooling and non-linear activation operations [41, 42]. CNNs were originally designed for image classification and object detection, and now are also adopted in image denoising.

Jain and Seung \cite{jain2009natural} demonstrated that convolutional neural networks (CNNs) can be used for image denoising and claimed that CNNs have achieved comparable or even superior performance than the MRF methods. Zhang et al. \cite{zhang2017beyond} proposed to incorporate residual learning and batch normalization learning strategies into very deep CNN for denoising. Mao et al. \cite{mao2016image} proposed to use skip-layer connection to symmetrically link convolutional and deconvolutional layers, which is able to train even deeper CNN architecture for denoising. Peng and Fang \cite{DBLP:journals/corr/LiuF17} proposed a wider CNN network which has relatively fewer layers but has larger size and number of filters in each layer. They claimed that for low-level vision tasks, the depth of the network is not the key, while the width of the architecture is more important. They state that for denoising tasks, deep learning denoising models learn prior pixel distribution information from original image and then use the learned filter banks to restore degrade images. Thus, the more concentrated convolutions to capture the prior image distribution from noisy images, the better the denoising performances.

\subsection{Residual learning}

 Residual learning is a technique to solve the gradient vanish problem \cite{he2016deep}. As the the number of layers increases, the training accuracy of CNN begins to decrease due to gradient vanishing in lower layers.   By constructing  residual units (i.e., identity shortcuts or skip connections) between a few layers, residual network learns a residual mapping which is much easier to train and prevent gradient vanish. With residual learning strategy, training extremely deep CNN become possible. He et al \cite{he2016deep} shows improved performance when using residual learning for image classification and object detection.

 There are several studies that incorporate residual learning for denoising tasks\cite{mao2016image,zhang2017beyond,DBLP:journals/corr/LiuF17}.
 In \cite{mao2016image}, they used Skip shortcuts to connect from convolutional feature maps to their corresponding deconvolutional feature maps every a few layers, which help ease back-propagation and reuse details. In \cite{zhang2017beyond}, Zhang et al. proposed DnCNN to using a mapping directly from an input observation to the corresponding reference observation.

\subsection{Batch normalization}

 Batch Normalization is another strategy to fasten the training process and improve the training accuracy. BN was designed to prevent internal covariance shift due to mini-batch stochastic gradient descent (SGD) which changes the distributions of internal non-linearity inputs during training. BN is motivated by the fact that data whitening process improves performance. First, BN normalizes the output of the bottom layer (Conv or ReLU), dimension-wise with zero mean and unit variance within a batch of training images; Second, BN optimally shifts and scales these normalized activations.

\section{Methods}

\subsection{Model}

In this section, we present the proposed ASLDLD for denoising ASL MRI images. There are generally two steps to train a CNN model for a specific task: 1) design network architecture and 2) training the network. First, we modify the WIN network to make it suitable to denoise ASL MRI images. We set the depth and width of the network and incorporated grey matter (GM) probability map as a regularizer because CBF map shows a similar image contrast to that of a grey matter map.Second, we adopt residual learning and batch normalization strategy to fasten and stabilize model training.

\subsection{Data}
ASL data were acquired from 280 subjects using a pseudo-continuous ASL sequence (40 control/labeled image pairs with labeling time = 1.48 sec, post-labeling delay = 1.5 sec, FOV=22 cm, matrix=64x64, TR/TE=4000/11 ms, 20 slices with a thickness of 5 mm plus 1 mm gap). Each pair of control/labeled image is subtracted using[xx] method to generate one CBF image. CBF images were calculated and spatially normalized into the MNI space using ASLtbx \cite{wang2008empirical,wang2013arterial} and SPM12.

To maximally show the benefit of ASLDLD, we took the mean of first 10 CBF images without smoothing (meanCBF-10) as the input image while used the mean of all 40 CBF images (meanCBF-40) with smoothing and adaptive outlier cleaning \cite{wang2013arterial} as the reference. The ASLDLD was trained with data from 240 subjects’ 3D CBF maps (input and reference). The remained 40 subjects were used as test samples. For each subject, we extract every 3 slice from slice 36 to slice 60 of a 3D CBF image. Thus, the number of total 2d CBF images for training are $240 \times 9 = 2160$.  The normalized CBF image size is 91 $\times$ 109  and we set the patch size as 16 $\times$ 16 with the stride of 4. The training data set are cropped into 984960 patches in total.

\subsubsection{Problem formulation}
The input of our ASLDLD is a noisy observation $y = x + n$, where $x$ is the latent clean image, $y$ is the noisy image and $n$ is the noise added to $x$. Rather than directly learn the original unreferenced mapping $T(y) \rightarrow x$,  the proposed ASLDLD has a skip connection from input-end to output-end to learn the residual mapping $T(y) \rightarrow -n$, then it has $x = y + T(y)$. In order to learn the weights $\Theta$ of the CNN, we minimize the Mean Squared Error (MSE) between the reference images $x_i$ and noisy input images $y_i$. Thus, the loss function of ASLDLD is
 $$ l(\Theta) = \frac{1}{2N}\sum^N_{i=1}||y_i + T(y_i;\Theta) -x_i ||^2$$.

Furthermore, grey matter (GM) probability map was incorporated as a regularizer because CBF map shows a similar image contrast to that of a grey matter map. To further encode GM prior information to the network, we add GM probability map $\gamma_i$ of each subject during the training. This step can be formulated as $||\hat{y}_i -\gamma_i||$. Thus the loss function is as follows:

$$l(\Theta) = \frac{1}{2N}\sum^N_{i=1}||y_i + T(y_i;\Theta) -x_i ||^2 + \alpha ||y_i -\gamma_i||$$.
where $\alpha$ is a hyperparameter and we set $\alpha$ to 0.1 in our case.

\subsubsection{Network architecture}

\begin{figure}[h]
  \centering
  \includegraphics[width=14cm]{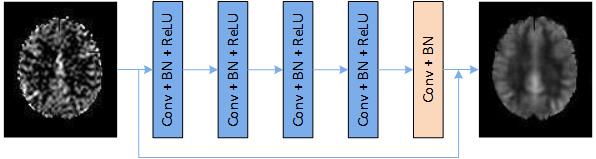}\\
  \caption{Figure 1: The architecture of the proposed ASLDLD network. }\label{network}
\end{figure}

The network architecture of ASLDLD, as shown in Figure \ref{network}, was based on WIN, a 5 layer Convolutional Neural Networks (CNNs) with wide structure. The “wide” structure here refers to the relatively larger size of filter (7x7) and the relatively larger number of filters (128) in each layer, while the number of layers (5) is fewer   compared to deep CNNs used in high-level tasks. The wider structure was used for two reasons: 1) larger filters can better utilize spatial correlation among neighboring voxels and 2) more filters are able to capture the pixel-level distribution information more effectively \cite{DBLP:journals/corr/LiuF17}.  The ASLDLD contains no signal pooling and no fully connected output layer as often used in regular CNNs. Each convolutional layer is followed by a ReLU, except for the last layer. The residual learning and batch normalization technique is further introduced to fasten and stabilize the training performance of ASLDLD.

\subsubsection{Implementation Details}

The network is trained end-to-end using ADAM \cite{kingma2014adam} with basic learning rate 0.1. Meantime, momentum 0.9, weight decay $1e-4$ and clip gradient 0.1 are also adopted to optimise training process.  We train the model using mini batches of size 64.  Caffe and MatConvNet packages are used to train the proposed ASLDLD models. All the experiments are running on a PC with Intel(R) Core(TM) i7-5820k CPU @3.30GHz and an Nvidia GeForce 980 Ti GPU. It takes about one day to train the ASLDLD on GPU.

\section{Results}
\subsection{Comparison with the state-of-the art}

\begin{figure}[h]
  \centering
  \includegraphics[width=10cm]{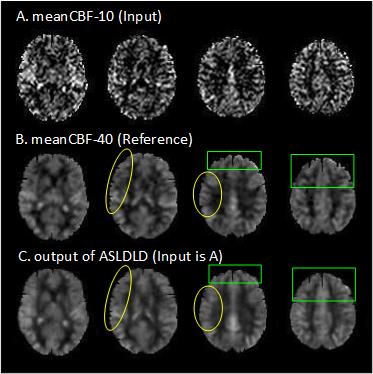}\\
  \caption{ASL CBF images. From top to bottom: Input images, non-DL-based denoising output images (reference images), ASLDLD output images. From left to right: Slices 40, 45, 50 and 55. }\label{cbf_re}
\end{figure}

ASLDLD was compared with current state-of-the art ASL denoising methods regarding the SNR of the resultant CBF images. Figure \ref{cbf_re} shows the resultant CBF maps from one representative subject. ASLDLD yielded superior performances to non-DL-based methods. Especially, ASLDLD recovered CBF signal in the air-brain boundaries as marked by the green boxes and signal loss due to partial volume effects as labeled by the yellow boxes. Slightly better texture was obtained by ASLDLD too. Figure 2 shows the SNR performance of different methods. SNR was calculated as the ratio between the mean signal of a GM region-of-interest (ROI) and the standard deviation of a white matter ROI. Compared to the non-DL-based methods, ASLDLD showed a 38.6\% SNR increase (p=1.14e-4).

In the experiment, we show ASLDLD achieve higher SNR than previous methods. Especially, it should be noted that ASLDLD only need 10 pairs of control labeling images rather than 40 pairs that needed for traditional methods. This significantly reduces the acquisition time of ASL MRI, which reduce the chance of head motions. This also help to reduce the noise introduced by head motions. Furthermore, because traditional methods generally involves a optimization process in the test stage while ASLDLD only has a feed-forward process, ASLDLD significantly reduce denoising time in the final test stages.

\begin{figure}[h]
  \centering
  \includegraphics[width=7cm]{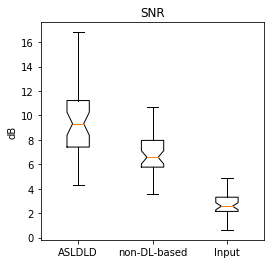}\\
  \caption{Fig. 2.: SNR of CBF test images using different methods. }\label{cbf_re}
\end{figure}

\section{Conclusion}

In this study, we showed that DL-based denoising can substantially improve ASL CBF SNR as well as the partial volume effects even only used the mean CBF map of 10 pairs of ASL control/label image acquisitions. In other words, ASLDLD can be utilized to significantly shorten the typical 5-6 mins acquisition time by 75\%, which would substantially reduce the chance of head motions, a big confound in ASL imaging. Besides, ASLDLD significantly reduce computing time in the test stage comparing to traditional methods.

\section*{References}
\bibliography{main}

\begin{thebibliography}{10}

\bibitem{aguirre2002experimental}
GK~Aguirre, JA~Detre, E~Zarahn, and DC~Alsop.
\newblock Experimental design and the relative sensitivity of bold and
  perfusion fmri.
\newblock {\em Neuroimage}, 15(3):488--500, 2002.

\bibitem{buades2005non}
Antoni Buades, Bartomeu Coll, and J-M Morel.
\newblock A non-local algorithm for image denoising.
\newblock In {\em Computer Vision and Pattern Recognition, 2005. CVPR 2005.
  IEEE Computer Society Conference on}, volume~2, pages 60--65. IEEE, 2005.

\bibitem{buades2008nonlocal}
Antoni Buades, Bartomeu Coll, and Jean-Michel Morel.
\newblock Nonlocal image and movie denoising.
\newblock {\em International journal of computer vision}, 76(2):123--139, 2008.

\bibitem{dabov2007image}
Kostadin Dabov, Alessandro Foi, Vladimir Katkovnik, and Karen Egiazarian.
\newblock Image denoising by sparse 3-d transform-domain collaborative
  filtering.
\newblock {\em IEEE Transactions on image processing}, 16(8):2080--2095, 2007.

\bibitem{detre1992perfusion}
John~A Detre, John~S Leigh, Donald~S Williams, and Alan~P Koretsky.
\newblock Perfusion imaging.
\newblock {\em Magnetic resonance in medicine}, 23(1):37--45, 1992.

\bibitem{detre2012applications}
John~A Detre, Hengyi Rao, Danny~JJ Wang, Yu~Fen Chen, and Ze~Wang.
\newblock Applications of arterial spin labeled mri in the brain.
\newblock {\em Journal of Magnetic Resonance Imaging}, 35(5):1026--1037, 2012.

\bibitem{dong2013nonlocally}
Weisheng Dong, Lei Zhang, Guangming Shi, and Xin Li.
\newblock Nonlocally centralized sparse representation for image restoration.
\newblock {\em IEEE Transactions on Image Processing}, 22(4):1620--1630, 2013.

\bibitem{elad2006image}
Michael Elad and Michal Aharon.
\newblock Image denoising via sparse and redundant representations over learned
  dictionaries.
\newblock {\em IEEE Transactions on Image processing}, 15(12):3736--3745, 2006.

\bibitem{he2016deep}
Kaiming He, Xiangyu Zhang, Shaoqing Ren, and Jian Sun.
\newblock Deep residual learning for image recognition.
\newblock In {\em Proceedings of the IEEE conference on computer vision and
  pattern recognition}, pages 770--778, 2016.

\bibitem{ioffe2015batch}
Sergey Ioffe and Christian Szegedy.
\newblock Batch normalization: Accelerating deep network training by reducing
  internal covariate shift.
\newblock In {\em International Conference on Machine Learning}, pages
  448--456, 2015.

\bibitem{jain2009natural}
Viren Jain and Sebastian Seung.
\newblock Natural image denoising with convolutional networks.
\newblock In {\em Advances in Neural Information Processing Systems}, pages
  769--776, 2009.

\bibitem{kingma2014adam}
Diederik Kingma and Jimmy Ba.
\newblock Adam: A method for stochastic optimization.
\newblock {\em arXiv preprint arXiv:1412.6980}, 2014.

\bibitem{lan2006efficient}
Xiangyang Lan, Stefan Roth, Daniel Huttenlocher, and Michael~J Black.
\newblock Efficient belief propagation with learned higher-order markov random
  fields.
\newblock In {\em European conference on computer vision}, pages 269--282.
  Springer, 2006.

\bibitem{li2009markov}
Stan~Z Li.
\newblock {\em Markov random field modeling in image analysis}.
\newblock Springer Science \& Business Media, 2009.

\bibitem{DBLP:journals/corr/LiuF17}
Peng Liu and Ruogu Fang.
\newblock Wide inference network for image denoising.
\newblock {\em CoRR}, abs/1707.05414, 2017.

\bibitem{liu2005signal}
Thomas~T Liu and Eric~C Wong.
\newblock A signal processing model for arterial spin labeling functional mri.
\newblock {\em Neuroimage}, 24(1):207--215, 2005.

\bibitem{lu2006detrimental}
Hanzhang Lu, Manus~J Donahue, and Peter van Zijl.
\newblock Detrimental effects of bold signal in arterial spin labeling fmri at
  high field strength.
\newblock {\em Magnetic resonance in medicine}, 56(3):546--552, 2006.

\bibitem{mairal2009non}
Julien Mairal, Francis Bach, Jean Ponce, Guillermo Sapiro, and Andrew
  Zisserman.
\newblock Non-local sparse models for image restoration.
\newblock In {\em Computer Vision, 2009 IEEE 12th International Conference on},
  pages 2272--2279. IEEE, 2009.

\bibitem{mao2016image}
Xiao-Jiao Mao, Chunhua Shen, and Yu-Bin Yang.
\newblock Image restoration using convolutional auto-encoders with symmetric
  skip connections.
\newblock {\em arXiv preprint arXiv:1606.08921}, 2016.

\bibitem{nair2010rectified}
Vinod Nair and Geoffrey~E Hinton.
\newblock Rectified linear units improve restricted boltzmann machines.
\newblock In {\em Proceedings of the 27th international conference on machine
  learning (ICML-10)}, pages 807--814, 2010.

\bibitem{roth2009fields}
Stefan Roth and Michael~J Black.
\newblock Fields of experts.
\newblock {\em International Journal of Computer Vision}, 82(2):205--229, 2009.

\bibitem{srivastava2014dropout}
Nitish Srivastava, Geoffrey~E Hinton, Alex Krizhevsky, Ilya Sutskever, and
  Ruslan Salakhutdinov.
\newblock Dropout: a simple way to prevent neural networks from overfitting.
\newblock {\em Journal of machine learning research}, 15(1):1929--1958, 2014.

\bibitem{wang2008empirical}
Ze~Wang, Geoffrey~K Aguirre, Hengyi Rao, Jiongjiong Wang, Mar{\'\i}a~A
  Fern{\'a}ndez-Seara, Anna~R Childress, and John~A Detre.
\newblock Empirical optimization of asl data analysis using an asl data
  processing toolbox: Asltbx.
\newblock {\em Magnetic resonance imaging}, 26(2):261--269, 2008.

\bibitem{wang2013arterial}
Ze~Wang, Sandhitsu~R Das, Sharon~X Xie, Steven~E Arnold, John~A Detre, David~A
  Wolk, Alzheimer's Disease~Neuroimaging Initiative, et~al.
\newblock Arterial spin labeled mri in prodromal alzheimer's disease: a
  multi-site study.
\newblock {\em NeuroImage: Clinical}, 2:630--636, 2013.

\bibitem{williams1992magnetic}
Donald~S Williams, John~A Detre, John~S Leigh, and Alan~P Koretsky.
\newblock Magnetic resonance imaging of perfusion using spin inversion of
  arterial water.
\newblock {\em Proceedings of the National Academy of Sciences},
  89(1):212--216, 1992.

\bibitem{wong1999potential}
EC~Wong.
\newblock Potential and pitfalls of arterial spin labeling based perfusion
  imaging techniques for mri.
\newblock {\em Functional MRI. Heidelberg: Springer}, pages 63--69, 1999.

\bibitem{zhang2017beyond}
Kai Zhang, Wangmeng Zuo, Yunjin Chen, Deyu Meng, and Lei Zhang.
\newblock Beyond a gaussian denoiser: Residual learning of deep cnn for image
  denoising.
\newblock {\em IEEE Transactions on Image Processing}, 2017.

\end{thebibliography}
\bibliographystyle{plain}
\end{document}